# OBSTACLE AVOIDANCE AND PATH FINDING FOR MOBILE ROBOT NAVIGATION

Poojith Kotikalapudi and Vinayak Elangovan

Division of Science and Engineering, Penn State Abington, PA, USA
pfk5148@psu.edu
Division of Science and Engineering, Penn State Abington, PA, USA
vue9@psu.edu

## *ABSTRACT*

This paper investigates different methods to detect obstacles ahead of a robot using a camera in the robot, an aerial camera, and an ultrasound sensor. Various efficient path finding methods are also explored for the robot to navigate to the target source. Single and multi-iteration angle-based navigation algorithms were developed. The theta-based path finding algorithms were compared with the Dijkstra's Algorithm and their performance were analyzed.

## 1. INTRODUCTION

After a disaster like a hurricane or a tornado has hit and left the area, often the roads are blocked by debris and people can be trapped in the rubble. Additionally, the areas are often littered with dangers that make it unsafe for humans to navigate. Dangers can include live wires in water or unstable foundations. An ideal approach is to use a robot that can autonomously navigate to a target location. A cohesive program is needed with efficient image processing, obstacle avoidance, and path finding algorithms to create a path between the robot and a target.

An autonomous mobile robot is a machine that operates in a partially unknown and unpredictable environment. Mostly, mobile robots are used for surveillance, inspection, and transportation tasks. The robot must operate safely, i.e. it must stay away from hazards such as obstacles or operating conditions dangerous to the robot itself, and it must pose no risk to humans in the vicinity of the robot. For any kind of mobile robots, navigation is a fundamental capability. One of the most challenging tasks for the mobile robot is to understand the information provided through various sensors, which will guide the robot in the environment and reach the destination by avoiding obstacles in the environment. One important task of a mobile robot intelligence program is environment perception and navigation. Environmental perception enables the robot to be aware of its environment.

There are two kinds of navigational environments for the robot to navigate. Completely known Environments and partially known environments. In the completely known environment, the robot knows complete information about all objects in the robot environment before navigation starts. The status of an obstacle is said to be static when its position or orientation are relative to a known and fixed origin that does not change with time. The status of the obstacles in the environment change with time may be in position or orientation or both according to its origin. In this paper, static obstacles are considered.

Robot path planning is the determination of how the robot navigates in the environment to reach its target. The path planning involves computing the collision-free path between two locations. Path planning takes a significant part of the computation time for many simulations, mainly in high time-dependent environments where most of the agents are moving. Path planning is typically performed on one agent at a time.

In general, there are two path planning techniques namely, global and local path planning. In global path planning the complete information about the environment is known, and obstacles should be static. In this approach, the algorithm generates a complete path from the start point to the destination point before the robot starts its motion. Local path planning is done while the robot is moving. The robot needs to change the path if there is a change in the environment. If there are no obstacles in the environment the path would be the straight line. The robot will head straight until it detects an obstacle. If it detects an obstacle, the robot will use path planning algorithm to find a feasible path to reach the target. In our research, the latter approach is employed.

## 2. RELATED WORK

The following section highlights some of the relevant search work carried out in robot navigation, path finding and obstacle avoidance.

### 2.1 Driver Assistance System based on Raspberry Pi [1]

This research paper describes how to navigate a robot when there are boundary lines present that can guide the robot. An edge detection is performed, and Hough line transform is applied to detect the lines on a road. These lines are then used to create a frame of reference for the robot to navigate.

### 2.2 Intelligent Survelliance Robot with Obstacle Avoidance Capabilities Using Neural Networks [2]

This paper explores how neural networks can be used to detect obstacles ahead and then avoid them appropriately. They used a combination of ultrasound sensors and camera. This method has shown great results to avoid obstacles to the right, left, and in the front of the robot.

### 2.3 A Comprehensive Study on PathFinding Techniques for Robotics and Video Games [3]

Various techniques of path finding algorithms that are being used within the realm of robotics and video games are investigated. This paper reports that using a traditional grid technique, it takes up considerable memory compared to hierarchical techniques that give more accurate representation near obstacles and less detail to large open fields where a lot of processing power is not needed.

### 2.4 Robot Navigation Control Based on Monocular Images: An Image Processing Algorithm for Obstacle Avoidance Decisions [4]

This paper highlights the issues of the dynamic environment in robot navigation. The paper details a 2-step approach in which they first use image segmentation to separate different parts of the image. A Balanced Histogram Thresholding was developed to find an optimal thresholding value that divides the image into a foreground and background. Edge detection is used to partition an image on abrupt changes in intensity between pixels. They also discuss how the Hough line transform method is used to highlight boundary lines.

### 2.5 Online Aerial Terrain Mapping for Ground Robot Navigation [5]

This research combines UAV (Unmanned aerial vehicle) and UGV (Unmanned ground vehicle) in a coherent system with user access from a ground station. The robots use GPS to track positions relative to each other. Processing is split among 4 computers, 2 being on the drone and the robot, and the other two as the base stations. One computer focuses on telemetry and the other as the mission planner and directly controls the drone. The program initiates by receiving images from the drone, creates an orthomoasic map which is stitched from all the images. The UAV creates a terrain map from which a path finding algorithm is used to create a path the UGV follows. The UGV does a lidar scan and creates an obstacle map and then send commands to motors to turn the robot wheels.

## 2.6  Shortest Path Finding and Tracking System Based on Dijkstra's Algorithm for Mobile Robot [6]

This research paper explores the Dijkstra's Algorithm specifically for mobile robot navigation. The research used a car-like robot with front-wheel steering. There were three different classifications of the shortest path algorithm which are single-source shortest path algorithm, single destination shortest path algorithm, and all-pairs shortest path algorithm.

## 2.7  Deep Learning using Rectified Linear Units (ReLU) [7]

This paper explored the performance of Rectified Linear Units with neural networks. They stated that certain neurons do not get activated properly and eventually die. This will affect the training of the algorithm and can often impede it. However, even with that drawback ReLU generally is seen as an effective method to prevent returning a negative value.

## 2.8  Image-Based Segmentation of Indoor Corridor Floors for a Mobile Robot [8]

A novel approach was proposed by combining three different parts of the image and using those visual cues to isolate the floor. The method proposed is extremely effective with the algorithm detecting 90% of the wall floor boundary. This algorithm is also optimized well for real time mobile robot navigation. A limitation has been stated if the floor is highly textured it can then become very difficult to detect. This can be a potential problem with carpets being potentially hard to detect.

## 2.9  Monocular Vision for Mobile Robot Localization and Autonomous Navigation [9]

This paper explored a method where a singular camera is used to navigate outdoor environments. This approach was compared to RTK GP. A 3d map was created to help the robot navigate. The paper states there has been difficulty updating the 3d map as the localization algorithm is not capable of a monitoring dynamic environment efficiently.

## 2.10  Obstacle Avoidance of Mobile Robot Based on HyperOmni Vision [10]

This research paper combines two approaches which are IDWA (Improved Dynamic Window Approach) and artificial potential field to avoid obstacles. The robot uses a concept of attraction and repulsion that would create a repulsive force when near an obstacle. This would cause the robot to move out of the way of the obstacle. The robot also establishes a dynamic window and predicts trajectories using the OmniHyper camera.

One approach for robot navigations is to employ an aerial camera to capture images over the targeted area and create a 2 D map to trace a path between the robot the target. A raspberry pi-controlled robot with a camera can be programmed using image processing techniques to identify obstacles and traverse around them. This research paper follows the above-mentioned approach for autonomous robot navigation. Various algorithms are developed in three main areas namely: robot navigation, path finding, and obstacle avoidance. Robot navigation focuses on interpreting the signals as well as processing the information that the robot will receive. The robot is programmed to detect the lines on a road and then calculates a path to navigate to the target. This was done using Hough line transform method. The next portion of the research paper highlights a different path finding techniques that can be used. A neural network model is developed and trained using YOLO object identification system to detect objects encountered in the navigation path. Data from an ultrasound sensor and imagery data from the robot camera are processed for obstacle detection. Combining the techniques of time of flight principle and ray tracing from the camera improves the efficacy of obstacle avoidance. These methods are built on a framework that uses a client-server protocol that allows communication between the robot and a workstation. The remaining part of this paper is organized as such: robot navigation, path finding, obstacle avoidance, a combination of systems, results, and conclusions followed by references.

## 3. ROBOT NAVIGATION

In order for the robot to navigate the map and get to the target, a 10-foot by 10-foot square testbed was created with lines the robot needed to follow. These lines lead the robot to the target. A lane keep assist program was developed to run on the workstation and send signals to the raspberry pi robot.

### 3.1 Lane Keep Assist Algorithim

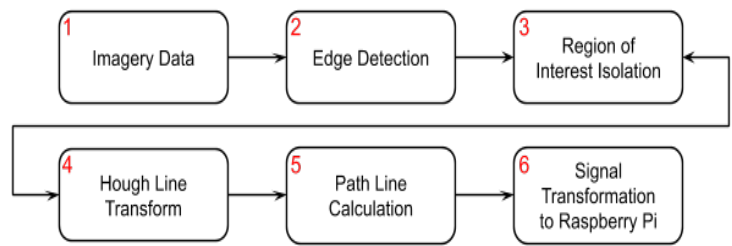

Figure 1: Lane Keep Assist Process

Figure 1 shows the step by step process in lane keep assist program. In step 1, the program received imagery data or a live stream video from the raspberry pi. After reading the image in step 2, the canny edge detection algorithm is applied to highlight the contours of the lines of the road. In step 3, a Region of Interest (RoI) is identified based on where the lines from the floor are in the image, and it was highlighted and isolated. This is done to reduce the processing time and is important especially for the Hough line transform method. RoI is altered based on changes in the environment. Factors to consider while calibrating the RoI is the width of the road, the height of the camera relative to the ground, and the focal length of the camera. In step 4, the Hough line transform method is used to identify the lines on the image. This returned the endpoints for each line on the floor to guide the robot. In step 5, the distance between the endpoints was calculated to get the midline of the road. Using the endpoints of the midline, a theta value was calculated for the robot to make necessary turns following the path. Figure 2 below shows the detection of the lane and path for the robot to navigate.

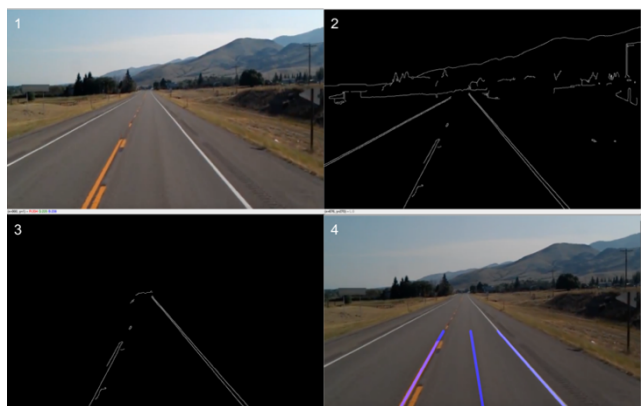

Figure 2: Lane Keep Assist Detections

## 3.2 Techniques of Angle Based Navigation

Multi iteration-based theta value:
In the first step, data from the previous iteration of the endpoints are saved; this has been done by using a FIFO data structure or a queue. In the second step, after the program passed the data values from the previous two iterations, the top endpoint and the bottom endpoint of the previous iteration and the top endpoint of the current iteration are used to calculate the theta value as shown in Figure 3. Also, note that the top endpoint has an x and y value, however, the program takes the value from the previous iteration for the top endpoint and writes it over the current iteration's y value for the top endpoint. This is done to reduce computing power.

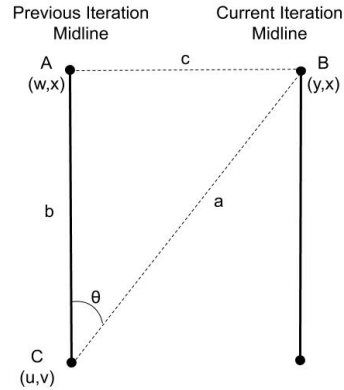
Figure 3: Diagram for Multi-Iteration theta values

The third step calculates the theta value as shown in equation-1 which combines the distance values and the law of cosines.

$$EQ\ 1: \theta = \cos^{-1}(\frac{a^2 + b^2 - c^2}{2ab}) * \frac{180}{2\pi}$$

$$EQ\ 2: a = \sqrt{(y - u)^2 + (x - v)^2}$$
$$EQ\ 3: b = \sqrt{(w - u)^2 + (x - v)^2}$$
$$EQ\ 4: c = w - y$$

In the fourth step, the slope of the line is calculated. If the slope is negative the theta value is negative. If the slope is positive, then the theta value is positive. Finally, the calculated theta value is sent back to the raspberry pi.

Single iteration-based theta value:
The program takes the endpoint values for the midline and creates a straight line from the x value of the bottom end point.

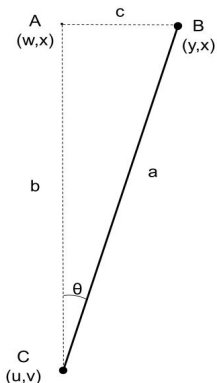
Figure 4: Diagram for Single-Iteration theta values

Equation 1 is used once again however different equations are used to find the variables c and b. The variable a however uses the same equation 2.

$$EQ\ 5: c = w - y$$
$$EQ\ 6: b = w - u$$

The third step is to find the slope needs to be calculated. If the slope is negative, then the theta value is negative. If the slope is positive the theta value is positive. The calculated theta value is sent to the raspberry pi to control the robot.

### 3.3 Advantages and Disadvantages of each technique

The advantages of the multi iteration-based theta value are that it gives the ability for the robot to correct itself when one or both of the lines are not detectable as it can use the previous values to gauge how far the robot has moved out of the line. The advantage of the single iteration-based theta value is that it is much more efficient and faster at returning a theta, something that is important when a robot is moving and needs to make quick decisions. However, the system was not accurate especially when the camera loses sight of one of the lines on the road. A future test is to use a combination of both ideas to retain the accuracy of the multi iteration-based theta value with the simplicity of the single iteration-based theta value. A proposed system can store previous iterations of midline while processing theta values in real-time by using the single iteration theta method. The two systems can verify each other with a margin for error.

### 3.4 Steering the robot with theta values

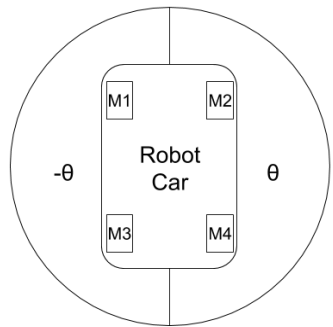

Figure 5: Diagram of the robot car with the right being all possible theta values and the left being all the negative theta values

Before a mathematical model is developed, an experiment was run to find out how much the robot turns in one second. To find the theta value the robot was placed on a line of tape in the testbed. The robot turned for 1 second. and the tape was placed where the robot turned. A protractor was used to measure by what degree the robot has turned. This was done 5 times to find the average of theta values. This returned the result that the robot turned 23 degrees every second. Therefore, a relationship can be stated that for every degree the robot needs to turn for 0.0435 seconds.

$$EQ\ 6: t = \frac{1}{23}|\theta|$$

An absolute value was used to prevent returning a negative time. After finding the seconds needed to turn, the robot then needs to find which way it will turn. A simple if-else statement is implemented where if the theta value was negative it will turn left but if the theta value was positive it will turn right.

## 4. PATH FINDING

Path finding is the method of formulating a path from one spot on the map to another spot on the map. Path finding will enable the program to determine the best path the robot should take to reach the target location. For the pathfinding algorithm originally, a camera was attached to moving platform on top of the area of interest. This moving platform could move in the x, y, and z-direction simulating the changing conditions of an aerial vehicle. Instead, a smartphone camera took a picture which is then sent by email to the workstation. Then the image is manually uploaded to the program. Numerous methods were used and tested keeping in mind of processing power and speed of the program.

### 4.1 Path finding algorithim using yolo object identification

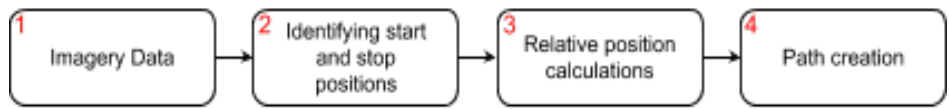

Figure 6: Block Diagram for theta-based path finding algorithm

In step 1, image data from the smartphone is manually inputted into the program. In step 2, to identify the start and stop positions the YOLO object identification was used. To identify the two objects a training program was run to train the machine learning and adjust the weights. The input nodes for the neural network were the pixels of the image. The algorithm learned by changing the weights of the hidden layers to get the desired outcome. Initially, pictures of the robot were taken to be used as the data set for the training algorithm. However, later on, it was found the algorithm was not identifying the robot consistently. The problem was found that there were not enough images of the robot. Instead, a JavaScript program was run that can download images of the raspberry pi. Then using the new dataset, a new set of weights was calculated that can be used to detect the raspberry pi on top of the robot. The same was then done to the target as well. To train the machine learning algorithm, GPU space was rented for free using Google Collab. The algorithm was trained for the maximum amount of time that google provided for free which was 12 hours. In step 3 once the start and end objects were identified the objects needed to be given a coordinate so a path can be formed.

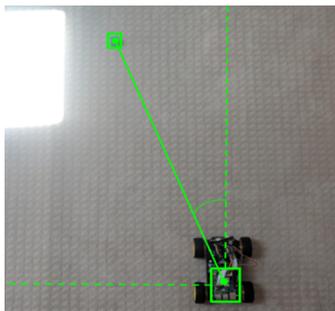

Figure 7: shows the theta angle calculation.

In step 4, once the coordinate values are given the path needs to be created for the robot. This was done by creating a straight line between the robot (start point) and the target (end point). Then using trigonometry, the angle theta was calculated to find how far the robot needs to turn to face the target. Once the robot has faced the target the robot went in a straight line until the robot encountered an obstacle which it will navigate around.

### 4.2 Advantages and Disadvantages

One of the big advantages of this method is simplicity. This method requires very little processing power and time is and is easy to code as well. The step which requires the most processing is the YOLO object identification. The process takes less time and processing power than alternatives like Dijkstra's Algorithm. Additionally, with Dijkstra's Algorithm, the value returned needed to be interpreted differently as it will not return a theta value which meant that an interpreter needed to be made to convert the values to GPIO signals for the robot to understand.

A major disadvantage of the program is its ability to not detect obstacles and create a path around it. This meant that the robot needed to perform the obstacle avoidance and there were no further redundancies. Also, if a hole was present in the ground it would be difficult for the robot to detect it compared to an overhead camera that can see it. Dijkstra's Algorithm with Yolo object identification, however, would be able to navigate around a hole.

## 5 OBSTACLE AVOIDANCE

### 5.1 Navigating with a ultrasonic sensor

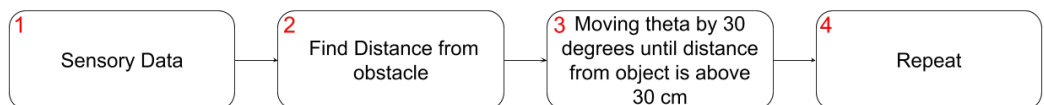

Figure 8: Block Diagram for obstacle avoidance program

In step 1, sensory data is received to the robot from the ultrasound sensor in the form of distance in cm from the obstacle. In step 2, the robot receives the distance from the robot and will be inputted into the program. In step 3, if the robot encounters an obstacle, the robot will keep moving forward until the obstacle is less than 30 cm away. The robot will then move 30 degrees to the right and then check the distance again to repeat the process.

If the obstacle is 30 cm away still, it will once again turn right by 30 degrees. If the obstacle is still 30 cm away from the robot after 90 degrees of motion. The robot will then turn left 120 degrees. The robot will continue to do the same to the left as well. This process will be repeated. Once the robot has navigated the obstacle the theta value at which the target is saved, and the robot will once again face the target and move straight.

## 6 HARDWARE

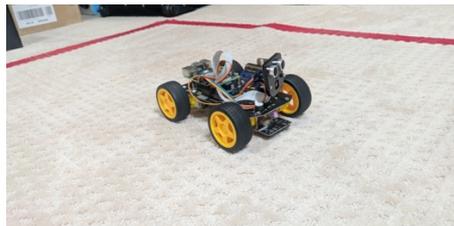

Figure 9: A sample of experimental set up.

The hardware used as an edge device is the raspberry pi 4b+. This particular model was chosen as it is the fastest raspberry pi. This speed helped greatly in making the program run faster. Another reason this was chosen is so the raspberry pi can act as an edge device that will be able to run its own programs. The robot chosen was Freenove 4WD Smart Car Kit for Raspberry Pi. The robot includes LEDs, speaker, lane detector, 4 motors for the wheels, an ultrasound sensor, and a camera. The lane detector was not used as the camera was used instead to detect lanes. A MacBook 15 inch was used as the workstation. The specification of the MacBook is an intel core i9 and 16 GB of ram. The overhead camera used was from a smartphone. To train any neural networks, google colab was used.

# 7 COMBINATION OF SYSTEMS

The program used yolo object identification to identify the target and the robot. Afterward, the program acquires the coordinate and calculates the theta between the two objects. The information is sent to the raspberry pi.

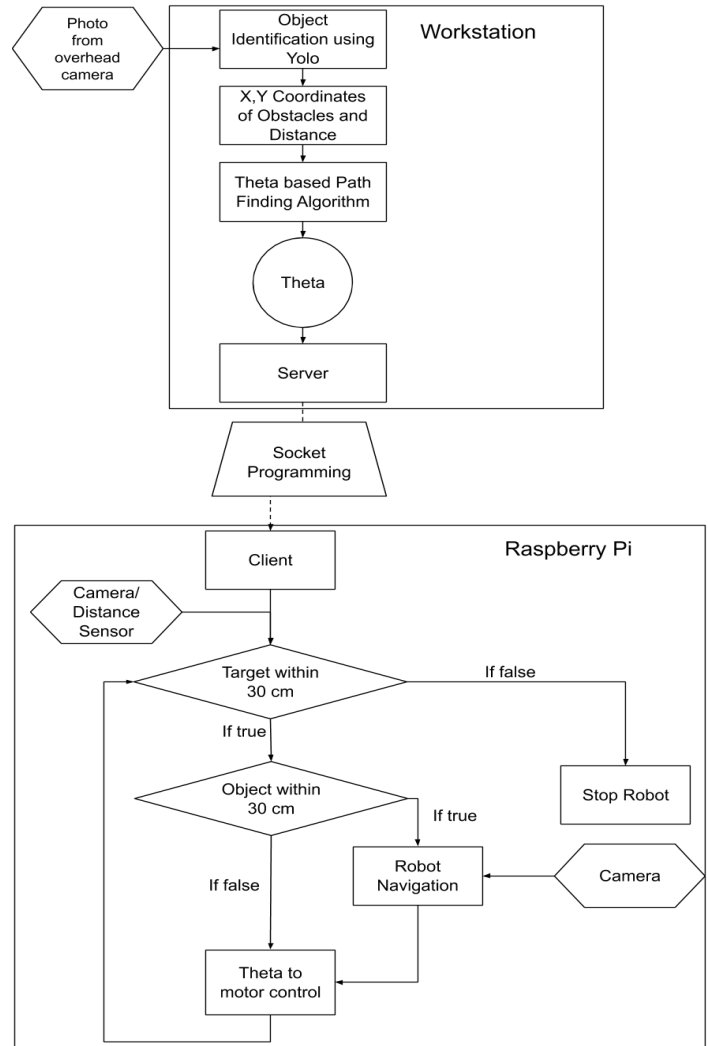

Figure 9: Block Diagram of the entire system

## 7.1 Socket Programming

The program sends the theta value from a server-class running on the workstation to the client that is running on the raspberry pi. The client first established a connection with the server on a specific port. The client will then listen to any data packages coming from the workstation. Once the client has received a package of data from the server, the program on the raspberry pi will start.

Once the theta value has been received, the robot will turn theta degrees to the left or right towards the target and start heading straight. The robot will check the distance sensor to see if an obstacle is detected within 30 cm of the robot. The camera will then be activated, and a picture is taken and further processed. The program will check for the target using Yolo object identification. If a target has not been detected the program returns a null and will navigate around the robot. The robot will then move towards the target and proceed straight, navigating around the obstacle when needed. Once it reaches the target the robot will first identify it using Yolo object identification. Afterward, the robot will stop and flash its LEDs and make a sound indicating that it has finished and reached the target.

## 8 RESULTS

Comparisons have been made between Dijkstra's Algorithm and theta-based path finding. Dijkstra's Algorithm on average takes 0.112 seconds while theta-based pathfinding algorithm took on average about $1.249 * 10^{-5}$ seconds. Dijkstra's Algorithm had 108 function calls compared to theta-based path finding had 9 function calls. The program was measured using cProfile which is a library in python. Theta based path finding had a lot less process running which contributed toward the program itself completing much faster. Theta based navigation is good at taking the best of multi computer processing and single computer processing. Since the robot only has to communicate with the workstation once to get the necessary theta value. Once the robot has that information it can navigate through the obstacles. This gives the robot the advantage of being able to utilize the resources of the workstation to find a clear path and also increases the range of the robot while not having to deal with the issue of an unreliable connection to the workstation. However, if there is an uneven surface (example: a pit) in the path, the robot would not detect it and would interpret it as an even surface ground. Another disadvantage is that due to the robot performing image processing using the raspberry pi, the reaction time of the robot slows down which can be potentially delay the rescue process in disaster ridden areas.

## 9 CONCLUSION

The research paper explored various techniques and how these techniques can be used together to form a coherent system. Lane keep assist and theta-based navigation system was developed and differences between single iteration and multi iteration lane keep assist were explored. It was found that with multi iteration the robot can remember its relative place even if one of the lines is not showing however with a single iteration-based system the program can be executed faster which lead to a faster response from the robot and follows the line more accurately. Path finding was another method investigated. A comparison was made between Dijkstra's Algorithm and a theta-based path finding algorithm where it was found that the theta-based path finding algorithm had a quicker run time which increased responsiveness time.

**Authors**

Poojith Kotikalapudi is an Undergraduate student at Penn State Abington. He is planning to major in Computer Science. He has a passion for robotics and coding. His major interests are building and programming robots to address real world problems.

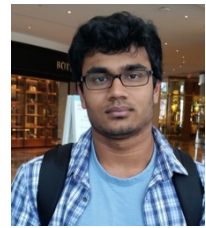

Dr. Vinayak Elangovan is an Assistant Professor of Computer Science at Penn State Abington. His research interest includes computer vision, machine vision, multi-sensor data fusion and activity sequence analysis with keen interest in software applications development and database management.

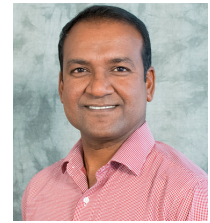